\documentclass[letterpaper, 10pt, conference]{ieeeconf}
\IEEEoverridecommandlockouts 
\usepackage{multicol}
\usepackage[bookmarks=true]{hyperref}

\usepackage{xparse} 
\usepackage{amsmath} 
\usepackage{amssymb}
\usepackage{amsfonts} 
\usepackage{dsfont}
\usepackage{graphicx}
\usepackage[caption=false]{subfig}
\usepackage{cite}
\usepackage{url}
\usepackage{booktabs}
\usepackage[para]{threeparttable}
\usepackage{hyperref}
\hypersetup{draft,colorlinks,linkcolor={blue},citecolor={green},urlcolor={blue}}

\usepackage{amsthm}
\usepackage{xcolor}
\usepackage{framed}\usepackage{xcolor,cancel}

\colorlet{shadecolor}{gray!10}

\let\oldthempfootnote\thempfootnote
\def\thempfootnote{\text{\oldthempfootnote}}
\usepackage[keeplastbox]{flushend}
\usepackage{color}
\usepackage{subfiles}
\usepackage[capitalize]{cleveref}
\IEEEtriggeratref{14}

\NewDocumentCommand\bbm{}{ \begin{bmatrix} }
\NewDocumentCommand\ebm{}{ \end{bmatrix} }
\NewDocumentCommand\Vector{m}{ \boldsymbol{\mathbf{#1}} }
\NewDocumentCommand\Matrix{m}{ \boldsymbol{\mathbf{#1}} }

\NewDocumentCommand\Norm{m}{\left\Vert#1\right\Vert }

\NewDocumentCommand\Real{}{ \mathbb{R} }

\NewDocumentCommand\LieGroupSE{m}{ \mathrm{SE}(#1) }

\NewDocumentCommand\Identity{}{ \Matrix{I} }

\NewDocumentCommand\CoordinateFrame{m}{ \underrightarrow{\Matrix{\mathcal{F}}}_{#1} }

\title{\Large\bf A Continuous-Time Approach for 3D Radar-to-Camera Extrinsic Calibration}

\author{Emmett Wise$^1$, Juraj Per\v{s}i\'{c}$^2$, Christopher Grebe$^1$, Ivan Petrovi\'{c}$^2$, and Jonathan Kelly$^{1,\dagger}$
\thanks{$^1$Emmett Wise, Christopher Grebe, and Jonathan Kelly are with the Space \& Terrestrial Autonomous Robotics Systems (STARS) Laboratory at the University of Toronto Institute for Aerospace Studies, Toronto, Canada. {\tt\footnotesize <firstname>.<lastname>@robotics.utias.utoronto.ca}}
\thanks{$^2$Juraj Per\v{s}i\'{c} and Ivan Petrovi\'{c} are with the Laboratory for Autonomous Systems and Mobile Robotics, University of Zagreb Faculty of Electrical Engineering and Computing, Croatia. {\tt\footnotesize <firstname>.<lastname>@fer.hr}}
\thanks{$^\dagger$Jonathan Kelly is a Vector Institute Faculty Affiliate. This research was supported in part by the Canada Research Chairs program.}}

\begin{document}
\maketitle

\begin{abstract}
Reliable operation in inclement weather is essential to the deployment of safe autonomous vehicles (AVs).
Robustness and reliability can be achieved by fusing data from the standard AV sensor suite (i.e., lidars, cameras) with \emph{weather robust} sensors, such as millimetre-wavelength radar. 
Critically, accurate sensor data fusion requires knowledge of the rigid-body transform between sensor pairs, which can be determined through the process of extrinsic calibration.
A number of extrinsic calibration algorithms have been designed for 2D (planar) radar sensors---however, recently-developed, low-cost 3D millimetre-wavelength radars are set to displace their 2D counterparts in many applications.
In this paper, we present a continuous-time 3D radar-to-camera extrinsic calibration algorithm that utilizes radar velocity measurements and, unlike the majority of existing techniques, does not require specialized radar retroreflectors to be present in the environment.
We derive the observability properties of our formulation and demonstrate the efficacy of our algorithm through synthetic and real-world experiments. 
\end{abstract}

\section{Introduction}
\label{sec:intro}

Safety is a paramount concern for autonomous vehicles (AVs) operating in human-centric environments (e.g., self-driving cars travelling on city streets).
To reduce the risk of failure and improve robustness, most AVs fuse data from multiple sensors on board.
The standard AV sensor suite typically includes cameras and lidar units; while these sensors are able to provide a high degree of situational awareness, they may fail to work reliably in inclement weather (e.g., heavy rain or snowfall). 
In turn, many AV sensor platforms incorporate 2D (planar) millimetre-wavelength radar units that are \emph{weather robust}---radar measurements are relatively immune to interference caused by precipitation, for example.

All radar sensors operate on the same basic principle: a low-frequency electromagnetic (EM) pulse is emitted from the radar antenna, reflects off of radar-opaque targets in the environment, and returns to the sensor.
By measuring the time of flight and phase of the return pulse, the radar is able to determine the azimuth, range, range-rate (velocity in the radial direction), and cross-section (reflectivity) of targets.
Low-frequency EM waves are able to pass through rain, snow, and other obscurants \cite{gourova_analysis_2017}. 
Although 2D radar has proven useful for many AV applications, the lack of complete 3D information limits its utility in many cases.

More recently, low-cost 3D radar sensors, such as the Texas Instruments AWR1843BOOST, have become available. 
Because of the additional information contained in 3D radar measurements (i.e., elevation), 3D radars are poised to replace 2D sensors in AV systems and in other applications.
To properly fuse 3D radar data with measurements from other AV sensors, however, knowledge of the rigid-body transform between the radar and the other sensors is required.
The process of determining the transform is known as extrinsic calibration. 
Often, extrinsic calibration is performed prior to deployment, in a laboratory or factory setting; the transform parameters are prone to change, however, due to material fatigue or user modifications.
Consequently, there is a need for methods to estimate the extrinsic calibration in the field.

\begin{figure}[t]
	\centering
	\includegraphics[width=0.95\columnwidth]{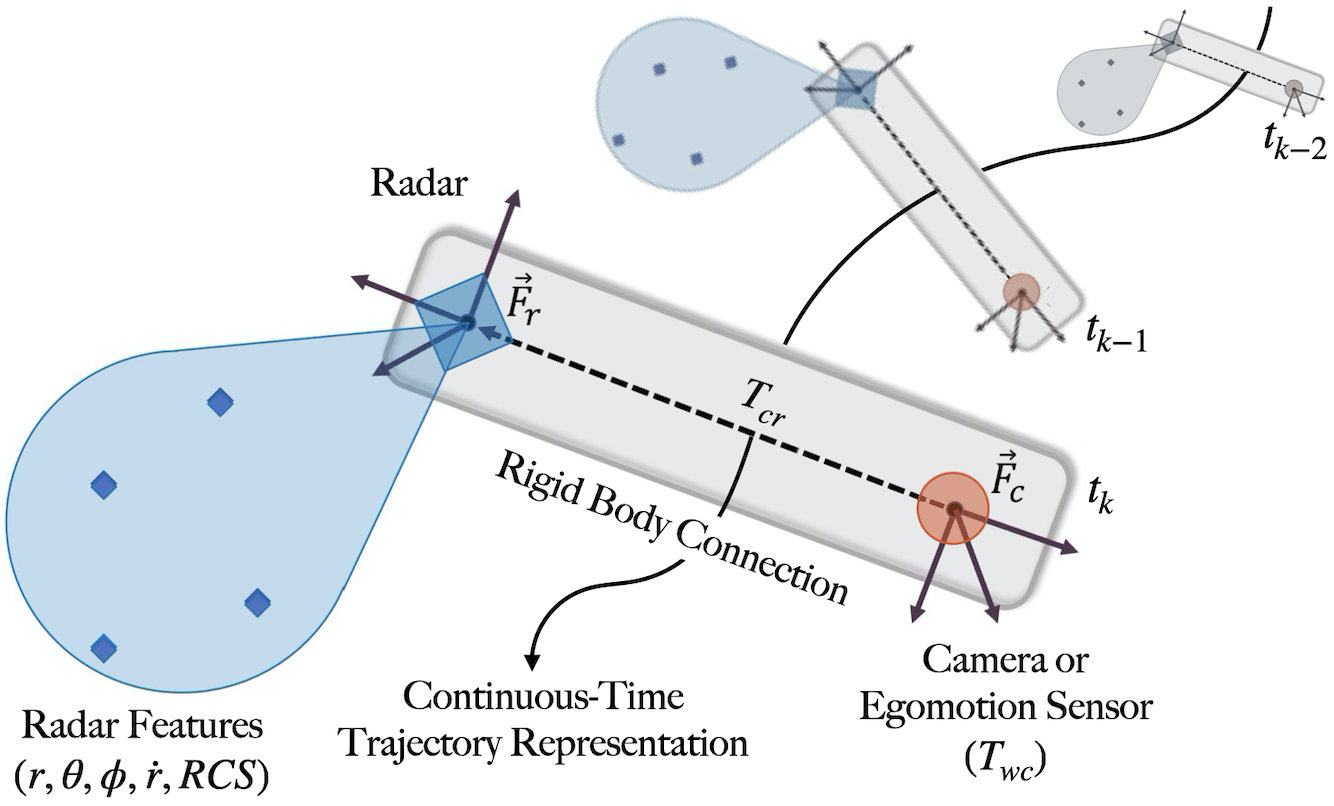}
	\caption{Depiction of the calibration problem. The radar measures the range, azimuth, elevation, range-rate, and reflectivity of objects in the environment. The camera (or egomotion sensor) measures its own pose change relative to a fixed reference frame. Our goal is to recover the rigid-body transform $\Matrix{T}_{cr}$ between the radar unit and the camera.}
	\label{fig:radarcal}
	\vspace{-5mm}
\end{figure}

Radar extrinsic calibration is challenging for several reasons. 
First, most radar measurement models assume that the EM pulse is reflected by one surface only.
In reality, there are often multipath reflections from several different surfaces. 
These multipath reflections create measurement outliers that can obscure or `drown out' the true reflection from a target.
Second, raw radar measurements have substantial jitter, which reduces measurement precision.
Finally, a radar pulse is a wave, and hence the exact point of reflection from a target can be ambiguous and/or inconsistent \cite{richards_principle_2010}.
The low precision and high outlier rate of radar measurements can degrade estimates of the extrinsic calibration.
To mitigate some of these issues, many existing calibration algorithms rely on specialized radar retroreflectors that are placed strategically in the environment.
Although this approach improves calibration, specialized retroreflectors are rarely available in the field during regular operation.

We overcome the challenges of radar extrinsic calibration by relying on the \emph{motion} of the sensor platform rather than on specific scene structure (see \cref{fig:radarcal}).
Work by Stahoviak has shown that the velocity of a 3D millimetre-wavelength (hereafter, mm-wave) radar sensor can be determined directly and without knowledge of the environment \cite{Stahoviak_Velocity_2019}.
By relying on velocity information provided by the 3D radar, instead of attempting to localize and track specific targets, we avoid many of the issues caused by noise, outliers, and jitter.
We focus on radar-to-camera extrinsic calibration---however, the method we describe is applicable to any complementary sensor that is able to estimate its egomotion (e.g., 3D lidar, GNSS/INS sensors, etc.).
We require only enough information for egomotion estimation and sufficient excitation of the system (see Section \ref{sec:deg_an}). 
In this paper we:

\begin{enumerate}
\item prove that extrinsic calibration for a 3D radar-camera pair is observable given sufficient excitation of the system;
\item describe the required motions necessary for proper calibration;
\item develop a continuous-time batch radar-to-monocular camera extrinsic calibration algorithm; and
\item verify the performance of our algorithm on synthetic data and through extensive real-world experiments.
\end{enumerate}

We provide one of the first methods for estimating the extrinsic calibration parameters between a 3D mm-wave radar and monocular camera without the use of radar retroreflectors. 
Although our goal is to build weather-robust navigation platforms, we focus on calibration under nominal conditions in the field (i.e., without adverse weather), since this is already a very difficult problem.

\section{Related Work} 
\label{sec:related_work}

A variety of mm-wave radar extrinsic calibration algorithms exist, which can roughly be grouped according to the sensor pair involved and the specific degrees of freedom that are calibrated.
Early extrinsic calibration algorithms for radar-camera sensor pairs considered 2D radar units only, either ignoring the 3D nature of radar measurements or constraining the positions of any retroreflectors to the radar measurement plane \cite{Sugimoto2004,Wang2011,kim_data_2014,kim_radar_2018}.
These algorithms operate by estimating the homography between the camera image plane and the radar measurement plane. 
Sugimoto et al.\ note in \cite{Sugimoto2004} that 2D radar units typically measure a maximum return when a retroreflector lies on the plane of zero elevation in the radar reference frame; the return intensity decreases for reflectors that lie above or below this plane.
The approach in \cite{Sugimoto2004} filters returns by intensity to ensure that only targets in the plane at zero elevation (relative to the radar frame) are used as part of the calibration process.

More recent algorithms estimate the rigid sensor-to-sensor transform by minimizing a `reprojection error': this is the error in the alignment of identifiable environmental structures or objects that appear within the fields of view of both sensors.
Kim et al.\ \cite{kim_comparative_2017} align hybrid visual-radar targets that can be easily identified in the camera and radar data, but assume that the radar measurements are constrained to the zero-elevation plane. 

The zero-elevation plane constraint is relaxed for certain `reprojection error' algorithms.
El Natour et al.\ estimate the radar-to-camera transform by intersecting backprojected camera rays with the `arcs' in 3D along which radar measurements must lie \cite{elnatour_radar_2015}.
Domhof et al.\ rely on a known visual target structure to convert camera measurements into  `pseudo-radar' measurements. The transform that best aligns the radar and pseudo-radar measurements then defines the extrinsic calibration \cite{Domhof2019_calibration}.
Per\v{s}i\'{c} et al.\ \cite{Persic2019_calibration} improve upon these methods by resolving the elevation ambiguity using target reflection intensity as a pseudo-measurement of the elevation angle.
Per\v{s}i\'{c} et al.\ \cite{Persic2019_calibration} also extend their approach to include 2D radar-to-lidar calibration.
The reprojection and homography methods are summarized and compared by Oh et al.\ in \cite{Oh2018_calibration}, where the authors conclude that the homography and reprojection methods have similar accuracy.

All of the algorithms described above require specialized retroreflective radar targets, but a small number of `targetless' or target-free extrinsic calibration algorithms for 2D mm-wave radar also exist.
Sch{\"{o}}ller et al.\ \cite{Knoll2019_calibration} use end-to-end deep learning to estimate the extrinsic rotation parameters that align vehicles (i.e., automobiles) detected in radar measurements and camera images.
However, the algorithm requires an external measurement of the translation parameters.
Per\v{s}i\'{c} et al.\ \cite{Persic2020} perform target-free, online pairwise extrinsic calibration of 2D radars, cameras, and lidar sensors by estimating the transform that aligns moving object trajectories.
This method assumes a priori knowledge of the translation parameters and only estimates yaw between the radar-camera and radar-lidar pairs.

Similar to our approach, Kellner et al.\ \cite{Kellner2015_calibration} use radar velocity measurements to estimate the yaw angle between a 2D radar sensor and a vehicle-mounted gyroscope, by relating the angular velocity of the gyroscope to the lateral velocity of the radar. 
This technique also requires a priori knowledge of the translation between the sensors.

In summary, the mm-wave radar calibration algorithms developed to date are generally limited by hardware constraints (i.e., an inability to resolve elevation reliably) or the need for specialized retroreflective targets, or suffer from high calibration parameter uncertainty due to a lack of true 3D information. 
We take advantage of the available elevation data in 3D radar measurements to estimate the instantaneous (3D) velocity of the radar unit. 
These data, in combination with pose estimates from a camera (or other egomotion sensors), allow us to determine the full sensor-to-sensor rigid-body transform without the need for specialized targets.

\section{Problem Formulation}
\label{sec:formulation}

\subsection{Notation}
\label{sec:notation}

Latin and Greek letters (e.g., $a$ and $\alpha$) represent scalar variables, while boldface lower and upper case letters (e.g., $\Vector{x}$ and $\Matrix{\Theta}$) represent vectors and matrices, respectively.
A parenthesized superscript pair, for example, $\Matrix{A}^{(i, j)}$, indicates the $i$th row and the $j$th column of the matrix $\Matrix{A}$. 
A three-dimensional reference frame is designated by $\CoordinateFrame{}$.
The translation vector from point $a$ (often a reference frame origin) to $b$, expressed in $\CoordinateFrame{a}$, is denoted by $\!\Vector{r}_a^{ba}$.
The translational velocity of point $b$ relative to point $a$, expressed in $\CoordinateFrame{c}$, is denoted by $\Vector{v}_c^{ba}$. 
The angular velocity of frame $\CoordinateFrame{a}$ relative to an inertial frame, expressed in $\CoordinateFrame{a}$, is denoted by $\!\Vector{\omega}_a$.

We denote rotation matrices by $\Matrix{R}$; for example, $\Matrix{R}_{ab} \in \mathrm{SO}(3)$ defines the rotation from $\CoordinateFrame{b}$ to $\CoordinateFrame{a}$. 
We reserve $\Matrix{T}$ for $\LieGroupSE3$ transform matrices; for example, $\Matrix{T}_{ab}$ is the homogeneous matrix that defines the rigid-body transform from frame $\CoordinateFrame{b}$ to $\CoordinateFrame{a}$.
These transforms are constructed using the split representation of $\LieGroupSE3$.
For example, the transform from frame $\CoordinateFrame{b}$ to $\CoordinateFrame{a}$ at time $t$ is,
\begin{equation}
\Matrix{T}_{ab}(t) = 
\bbm \Matrix{R}_{ab}(t) & \Vector{r}_a^{ba}(t) \\ \Vector{0}^T & 1 \ebm,
\end{equation}
where the transform is split into a rotation matrix, $\Matrix{R}_{ab}(t) \in \mathrm{SO}(3)$, and translation vector, $\Vector{r}_a^{ba}(t) \in \Real^3$.
The unary operator $^\wedge$ acts on $\Vector{r} \in \Real^3$ to produce a skew-symmetric matrix such that $\Vector{r}^\wedge\Vector{s}$ is equivalent to the cross product $\Vector{r} \times \Vector{s}$.

\subsection{Sensor Measurements}
\label{subsec:measurements}

We consider three reference frames: frame $\CoordinateFrame{w}$ is an (approximate) inertial frame attached to the surface of the Earth, while $\CoordinateFrame{r}$ is the reference frame of the radar sensor, and $\CoordinateFrame{c}$ is the reference frame of the camera (or other egomotion sensor).
The radar unit measures the velocity of the sensor in $\CoordinateFrame{r}$ relative to $\CoordinateFrame{w}$, expressed in $\CoordinateFrame{r}$ at an instant in time, $t$, 
\begin{equation} 
\label{eq:radar-vel}
\Vector{v}_r^{rw}(t) = \Matrix{R}_{wr}(t)^T\,
\frac{\partial\,\Vector{{r}}_w^{rw}(t)}{\partial t},
\end{equation}
where we use the partial derivative notation to indicate that the radar position also depends upon the parameters of our B-spline trajectory representation (see Section \ref{subsec:trajectory}).

Assuming that a series of three or more (known) 3D landmarks are visible in frame $\CoordinateFrame{w}$, the camera is able to measure its pose at time $t$ relative to $\CoordinateFrame{w}$,
\begin{equation} 
\label{eq:pose}
\Matrix{T}_{cw}(t) = \Matrix{T}_{cr}\Matrix{T}_{wr}^{-1}(t),
\end{equation}
where $\Matrix{T}_{wr}(t)$ is the homogeneous pose matrix of the radar in the inertial frame at time $t$ and $\Matrix{T}_{cr}$ is the homogeneous matrix that defines the (constant but unknown) radar-to-camera transform.
If the metric positions of the landmarks are not known, the camera translation can only be determined up to an unknown scale factor.

\subsection{Continuous-Time Trajectory Representation}
\label{subsec:trajectory}

We use a continuous-time representation of the sensor platform trajectory in our problem formulation.
The continuous-time representation is advantageous because it allows measurements to be made at arbitrary time instants; since the radar and the camera operate at different rates and are not hardware synchronized, the relationship between their measurement times is not fixed.
There are multiple possible ways to parameterize trajectories in continuous time \cite{Sommer_Efficient_2019,Furgale_Continuous_2015,barfoot2017state}.
We choose the cumulative B-spline representation on Lie groups developed by Sommer et al.\ in \cite{Sommer_Efficient_2019}.
Below, we very briefly review this representation, and refer the reader to \cite{Sommer_Efficient_2019} for more details.

B-splines are functions of one continuous parameter (e.g time) and a finite set of control points (or \emph{knots}); for brevity, we restrict our example here to points $\{\Vector{p}_0, \dots, \Vector{p}_N \mid \Vector{p}_i \in \Real^d\}$.
The order $k$ of the spline determines the number of control points that are required to evaluate the spline at time $t$.
In a uniformly spaced B-spline, each control point is assigned a time $t_i = t_0 + i \Delta t$, where $t_0$ is the start of the spline and $\Delta t$ is the time between control points.
Given a B-spline of length $N$ and order $k$, the end of the spline is $t_{N - k + 1}$.
 
Given a time $t$, a normalized time $u = \frac{t - t_i}{t_{i+1} - t_i}$ can be defined, where $t_i$ is the time assigned to control point $\Vector{p}_i$ and $t_i \leq t < t_{i+1}$.
The B-spline function evaluated at normalized time $u$ is
\begin{equation} \label{eq:rd_spline}
\Vector{p}(u) = \bbm \Vector{p}_i & \Vector{d}_1^i & \dots & \Vector{d}_{k-1}^i \ebm\tilde{\Matrix{M}}_{k}\Vector{u},
\end{equation}
where $\Vector{u}^T = [1 \; u \; u^2 \;\dots\; u^{k-1}]$ and $\Vector{d}_j^i = \Vector{p}_{i+j} - \Vector{p}_{i+j-1}$.
The matrix $\tilde{\Matrix{M}}_{k}$ is a $k\times k$ \emph{mixing matrix}. The elements of the mixing matrix are a function of the spline order $k$ and are defined by 
\begin{align} \label{eq:mix}
\tilde{m}^{(a,n)}_{k} &= \sum_{s=a}^{k-1}m^{(s,n)}_{k}, \\
\begin{split}
m^{(s,n)}_{k} &= \frac{C^n_{k-1}}{(k-1)!} \sum_{l=s}^{k-1}(-1)^{l-s}C_k^{l-s}(k-1-l)^{k-1-n}\\
& a,s,n \in \{0, \dots, k-1 \}.
\end{split}
\end{align}
The scalar $C^i_{j} = \frac{j!}{i!(j-i)!}$ is a binomial coefficient.
This B-splines definition can be simplified by defining $\lambda_j(u) = \tilde{\Matrix{M}}_{k}\Vector{u}$, which results in 
\begin{equation}
\Vector{p}(u) = \Vector{p}_i + \sum_{j=1}^{k-1}\lambda_j(u)\Vector{d}^i_{j}.
\end{equation}

This B-spline representation is a convenient way to describe smooth rigid-body trajectories in continuous time.
Our development above is for splines on a vector space, but B-splines can also be defined over Lie groups, including the group $\mathrm{SO}(3)$ of rotations,
\begin{equation}
\Matrix{R}(u) = \Matrix{R}_i\prod_{j=1}^{k-1}\exp(\lambda_j(u)\Vector{\phi}_j^i),
\end{equation}
where $\Matrix{R}_i$ is a control point of the rotation spline and $\Vector{\phi}^i_j = \log(\Matrix{R}_{i+j-1}^T\Matrix{R}_{i+j})$. 
The operators $\exp$ and $\log$ map from the Lie algebra $ \mathfrak{so}(3)$ to $\mathrm{SO}(3)$ and vice versa, respectively \cite{barfoot2017state}. 

\subsection{Optimization Problem}

The error equation for the radar velocity is
\begin{equation} 
\label{eq:radar-vel-error}
\begin{split}
\Vector{e}_v(t) &= \Vector{v}_r^{rw}(t) - \Matrix{R}_{wr}(t)^T\frac{\partial\,\Vector{r}_w^{rw}(t)}{\partial t} + \Vector{n}_v, \\
\Vector{n}_v &\sim \mathcal{N}(0, \Matrix{\Sigma}_v(t)),
\end{split} 
\end{equation}
where $\Matrix{R}_{wr}(t)$ and $\Vector{r}_w^{rw}(t)$ are the split spline representation of $\Matrix{T}_{wr}(t)$ with control points $\{\Matrix{R}_0, \dots, \Matrix{R}_N \mid \Matrix{R}_i \in \mathrm{SO}(3)\}$ and $\{\Vector{p}_0, \dots, \Vector{p}_N \mid \Vector{p}_i \in \Real^3\}$.
The vector $\Vector{v}_r^{rw}(t)$ is the measured radar velocity at time $t$.
The error equation for the camera measurements is
\begin{align} 
\label{eq:pose-error}
\Matrix{T}_{err}(t) &= \Matrix{T}_{cw}(t)\Matrix{T}_{wr}(t)\Matrix{T}_{cr}^{-1} \\
\Vector{e}_p(t) &= \bbm \Vector{r}_{err}(t) \\ \Vector{\phi}_{err}(t) \ebm +  \Vector{n}_p, \, \Vector{n}_p \sim \mathcal{N}(0, \Matrix{\Sigma}_p(t)) \\
\Vector{\phi}_{err}(t) &=  \log(\Matrix{R}_{err}(t)),
\end{align}
where $\Vector{r}_{err}(t)$ and $\Matrix{R}_{err}(t)$ are the $\Real^3$ and $\mathrm{SO}(3)$ elements of $\Matrix{T}_{err}(t)$. 
The set of parameters, $\Vector{x}$, that we wish to estimate are the control points of the split representation of $\Matrix{T}_{wr}(t)$ and the extrinsic calibration parameters in $\Matrix{T}_{cr}$,
\begin{equation}
\Vector{x} = \{\begin{matrix} \Vector{p}_0, & \dots, & \Vector{p}_N, & \Matrix{R}_0, & \dots, & \Matrix{R}_N, & \Matrix{R}_{cr}, & \Vector{r}_c^{rc} \end{matrix}\}.
\end{equation}
Our optimization problem is then to find $\Vector{x}^*$ that minimizes the following cost function:
\begin{align} 
\label{eq:opt-prob}
\begin{split}
\mathcal{J}(\Vector{x}) = & \sum_{i=1}^l \Vector{e}_v^T(t_i)\Matrix{\Sigma}_v^{-1}(t_i)\Vector{e}_v(t_i) \\ & + \sum_{j=1}^m \Vector{e}_p^T(t_j)\Matrix{\Sigma}_p^{-1}(t_j)\Vector{e}_p(t_j),
\end{split}
\end{align}
where $l$ and $m$ are, respectively, the number of radar velocity measurements and camera pose measurements.

\subsection{Implementation Details}

Our approach to estimate the velocity of the radar unit involves finding the velocity vector that best fits a series of measured range-rate vectors. 
To do so, we use an algorithm and software package developed by Stahoviak et al. called `Goggles' \cite{Stahoviak_Velocity_2019}.\footnote{Available at \url{https://github.com/cstahoviak/goggles}}
The Goggles algorithm applies MLESAC to find an inlier set of radar velocity measurements.
The final velocity estimate is calculated using orthogonal distance regression on this inlier set of velocities.

We solve the full batch nonlinear optimization problem to determine the extrinsic parameters using the Levenberg-Marquardt implementation available in the Ceres solver \cite{ceres-solver}. 
Ceres' auto-differentiation capability is applied to calculate the Jacobians of the error equations.
To manipulate the B-splines, we rely on the library from Sommer et al.\ \cite{Sommer_Efficient_2019}.\footnote{Available at \url{https://gitlab.com/VladyslavUsenko/basalt-headers.git}}
Our translation and rotation splines have a spline order of $k = 4$. 

\section{Observability Analysis}
\label{sec:observability}

In order to estimate the calibration parameters, the system must be observable (or, equivalently for our batch formulation, identifiable).
In Section \ref{sec:obs_an}, we make use of the observability rank condition criterion defined by Hermann and Krener \cite{hermannNonlinearControllabilityObservability1977} to prove that the calibration and scale estimation problem is observable.
It is well known that, in the absence of metric distance information, absolute scale cannot be recovered from monocular camera measurements alone \cite{chiuso2002_structure}.
We show below that, given radar velocity data, it is possible to identify both the calibration parameters and the visual scale factor \emph{without} knowledge of the (metric) distances between visual landmarks.
It follows that radar-to-camera calibration, in the general case, does not require a specialized camera calibration target (or any other external source of scale information).
We are concerned with the following set of parameters: 
\begin{equation}
\Vector{x} = \{\begin{matrix} \Vector{r}_c^{rc}, & \Matrix{R}_{cr}, & \alpha \end{matrix}\},
\end{equation}
where $\alpha$ is the unknown scale factor that appears in the camera pose measurement.
A brief degeneracy analysis of the calibration problem, which identifies conditions that result in a loss of observability, is provided in Section \ref{sec:deg_an}.

\subsection{Observability of Radar-to-Camera Extrinsic Calibration} 
\label{sec:obs_an}

We follow an approach similar to that in \cite{2014_Li_Online} and note that the (scaled) linear and angular velocities of the camera can be determined by taking the time derivatives of the camera pose measurements.
Also, Stahoviak has shown that the 3D velocity of the radar (in the radar frame) can be recovered from three non-coplanar range-rate measurements \cite{Stahoviak_Velocity_2019}.
These quantities can be related through rigid-body kinematics,
\begin{equation}
\Vector{h}_i = \alpha\Vector{v}_c^{cw} = \alpha (\Matrix{R}_{cr}\Vector{v}_r^{rw} - \Vector{\omega}_c^{\wedge}\Vector{r}_c^{rc}),
\end{equation}
where $\Vector{h}_i$ is the scaled linear velocity of the camera and $\Vector{\omega}_{c}$ is the angular velocity of the camera, both relative to the camera frame.
To decrease the notational burden going forward, we drop the superscripts and subscripts defining the velocities and extrinsic transform parameters.
The gradient of the zeroth-order Lie derivative of the $i$\hspace{0.05em}th measurement is
\begin{equation} \label{eqn:obs_meas}
\nabla_{\Vector{x}}L_0\Vector{h}_i = \bbm -\alpha\Vector{\omega}_i^\wedge & -\alpha(\Matrix{R}\Vector{v}_i)^\wedge\Matrix{J} & \Matrix{R}\Vector{v}_i - \Vector{\omega}_i^{\wedge}\Vector{r} \ebm,
\end{equation}
where $\Matrix{J}$ is the Lie algebra left Jacobian of $\Matrix{R}_{cr}$ \cite{barfoot2017state}.
Since the parameters of interest are constant with respect to time, we are able to stack the gradients of several Lie derivatives (at different points times) to form the observability matrix,
\begin{equation}
\Matrix{O} = \bbm \nabla_{\Vector{x}}L_0\Vector{h}_1 \\ \nabla_{\Vector{x}}L_0\Vector{h}_2 \\ \nabla_{\Vector{x}}L_0\Vector{h}_3 \ebm,
\end{equation}
which has full column rank when three or more sets of measurements are available (we omit the full proof for brevity).
We note that the analysis is simplified by considering the measurement equation only, and at different points in time. 
However, it is also possible to show that the system is instantaneously locally weakly observable when the sensor platform undergoes both linear and angular accelerations (again, we omit this proof due to space).

\subsection{Degeneracy Analysis} 
\label{sec:deg_an}

The conditions under which a loss of observability (identifiability) may occur can be determined by examining the nullspace of the observability matrix. In this section, we consider the scale parameter to be known, which removes the last column of the matrix defined by Eq. \ref{eqn:obs_meas}---in turn, only two sets of measurements are required. The nullspace of $\nabla_{\Vector{x}}L_0\Vector{h}_i$ contains the vectors
\begin{equation} 
\label{eqn:null}
\Matrix{U}_i = \bbm \Vector{\omega}_i & \Vector{0} & (\Identity - \frac{\Vector{\omega}_i\Vector{\omega}_i^T}{\Norm{\Vector{\omega}_i}^2})\Matrix{R}\Vector{v}_i \\[3mm]
\Vector{0}  & \Matrix{J}^{-1}\Matrix{R}\Vector{v}_i & (\Identity - \frac{\Matrix{J}^{-1}\Matrix{R}\Vector{v}_i(\Matrix{J}^{-1}\Matrix{R}\Vector{v}_i)^T}{\Norm{\Matrix{J}^{-1}\Matrix{R}\Vector{v}_i}^2})\Matrix{J}^{-1}\Vector{\omega}_i \ebm,
\end{equation}
where each column of $\Matrix{U}_{i}$ defines one null vector. To ensure that the stacked observability matrix formed from $\nabla_{\Vector{x}}L_0\Vector{h}_1$ and $\nabla_{\Vector{x}}L_0\Vector{h}_2$ has full column rank (i.e., that the nullspace contains the zero vector only), the following constraints must be satisfied, at minimum:
\begin{equation} 
\label{eqn:constraints}
\begin{aligned}
\Vector{\omega}_2\times\Vector{\omega}_1 &\neq 0,\\
\Vector{v}_2\times\Vector{v}_1 &\neq 0.
\end{aligned} 
\end{equation}
The constraints defined by Eq. \ref{eqn:constraints} show that the system must rotate about and translate along two non-collinear axes at different points in time.
The rotation constraint is expected because our problem is similar to the one defined by Brookshire and Teller in \cite{brookshire2013extrinsic}. 
However, the angular velocity of the radar unit cannot be measured directly, which leads to the second excitation requirement. 
Additional constraints can be generated from the third column of Eq. \ref{eqn:null}, but these motions are more difficult to characterize; we posit, based on our experiments, that these constraints are less likely to be violated in practice.

\section{Experiments and Results} 
\label{sec:experiments}

In general, our algorithm can be applied to any 3D radar and egomotion sensor pair, but our experimental focus is on 3D radar-to-monocular camera  extrinsic calibration.
For convenience, in this work, we estimate the camera pose relative to a $12 \times 10$ planar checkerboard calibration target of known size. 
However, as shown in Section \ref{sec:observability}, knowledge of metric scale is not required---the camera must simply view a sufficient number of features (three or more) that lie in a general configuration in the environment.

Below, we present a series of synthetic and real world calibration experiments to evaluate the performance of our algorithm.
In Section \ref{sec:synth}, we empirically analyze the sensitivity of the algorithm to measurement noise when applied to synthetic data.
In Section \ref{sec:real}, we demonstrate that our approach improves upon hand-measured calibration and compares favourably with the algorithm of Per\v{s}i\'{c} et al.\ \cite{persic_spatiotemporal_2019}, although our approach does not require specialized radar retroreflectors. 

\subsection{Synthetic Data} 
\label{sec:synth}

\begin{figure}[t]
	\centering
	\includegraphics[scale = 0.743]{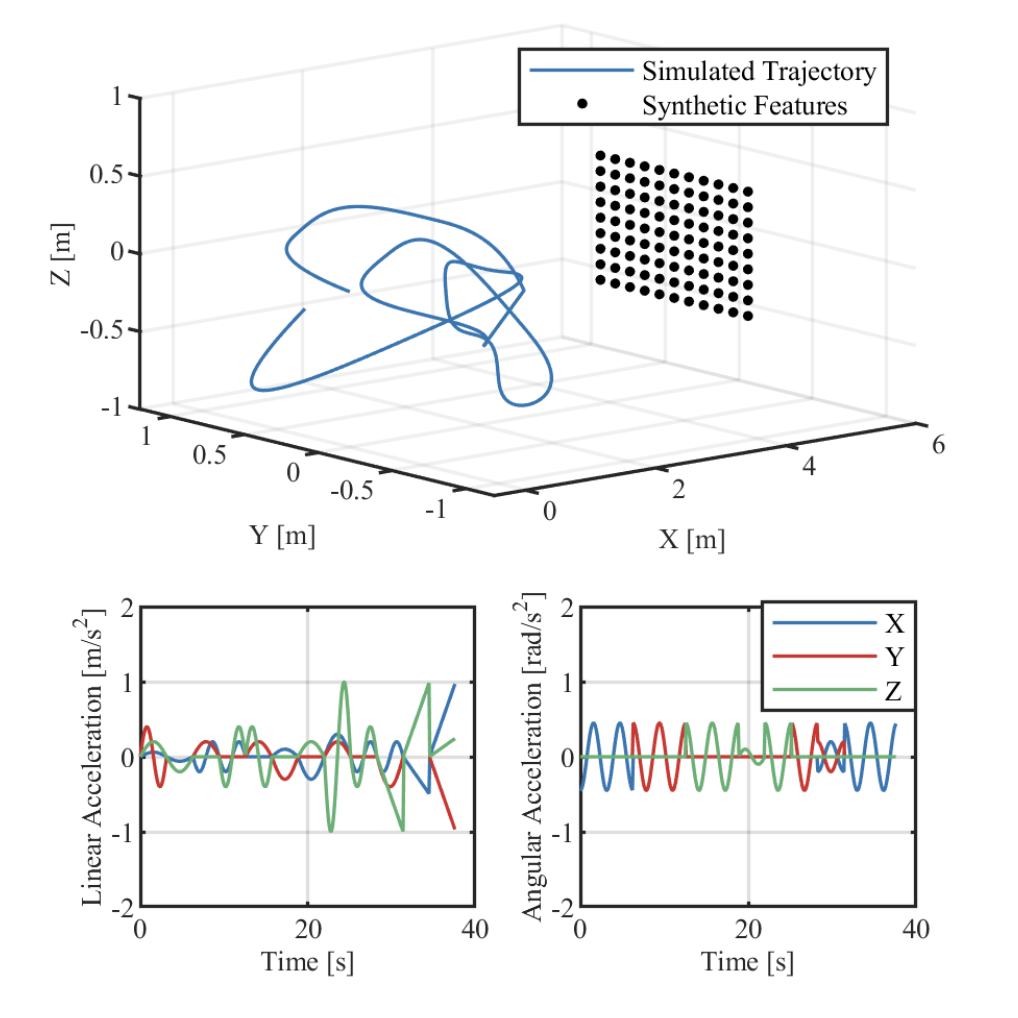}
	\vspace{-5mm}
	\caption{Experimental setup for our simulation studies. 
	The calibration rig rotates while moving along the blue trajectory.
	The black dots represent the internal corners of a 12-by-10 checkerboard with squares that are $9.9$ cm by $9.9$ cm in size, the same as those of our physical checkerboard.}
	\label{fig:synth_env}
	\vspace{-5mm}
\end{figure}

Our simulation environment is shown in \cref{fig:synth_env}.
In order to ensure sufficient excitation of the system, the sensor platform trajectory has non-zero linear and angular acceleration about all three axes in the radar sensor frame; see the bottom of \cref{fig:synth_env}.
We added zero-mean Gaussian noise to each radar and camera measurement, with magnitudes similar to the noise levels identified in our real-world experiments.

\begin{figure}[t]
	\centering
	\includegraphics[scale=0.8]{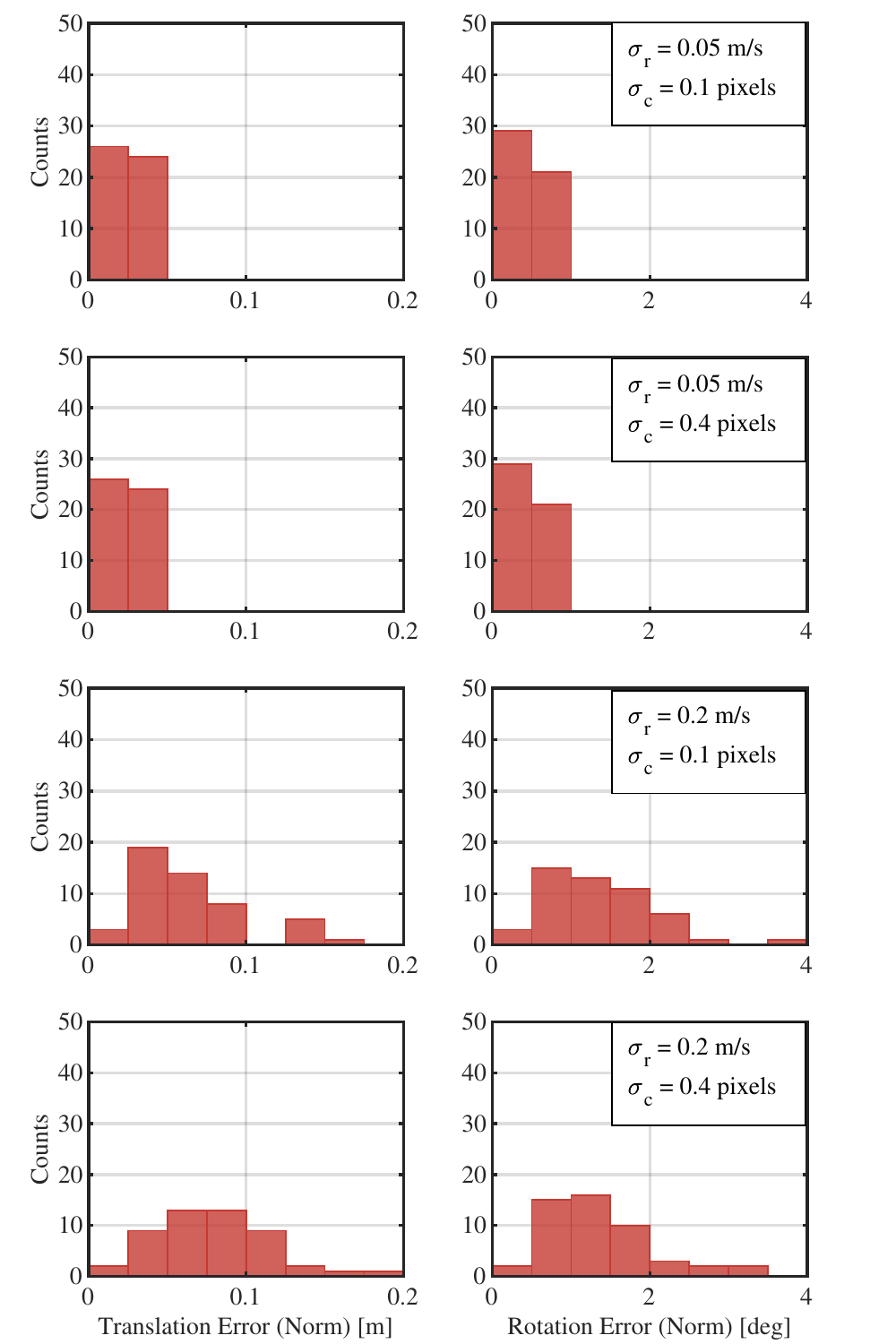}
	\caption{Left: histograms of translation error norm between estimated and ground truth calibration parameters for different amounts of simulated radar velocity and image pixel noise. 
	Right: histograms of rotation errors. 
	The rotation error is the magnitude of the angle that aligns the estimated and true radar frames. 
	For each noise combination, 50 test cases were run.  
}
	\label{fig:synth_result}
	\vspace{-0.6cm}
\end{figure}

Simulation results show that our algorithm is accurate in the low-noise regime, but that the performance degrades as the amount of noise in the radar velocity measurements increases (see Figure \ref{fig:synth_result}).
We found that the average standard deviations of our real-world radar velocity estimates were $0.03$, $0.06$, and $0.1$ m/s in the $x$, $y$, and $z$ directions, respectively. 
As a result, our noisiest simulation experiment represents a worst-case calibration scenario, because the experiment uses twice the amount of noise as found in our true radar velocity data. 
Overall, the proposed calibration algorithm shows robustness to significant noise---we are able to successfully calibrate in all of our trials despite very large worst-case noise levels.

\subsection{Real-World Experiments} 
\label{sec:real}

\begin{figure}[t]
	\centering
	\includegraphics[width=0.98\columnwidth]{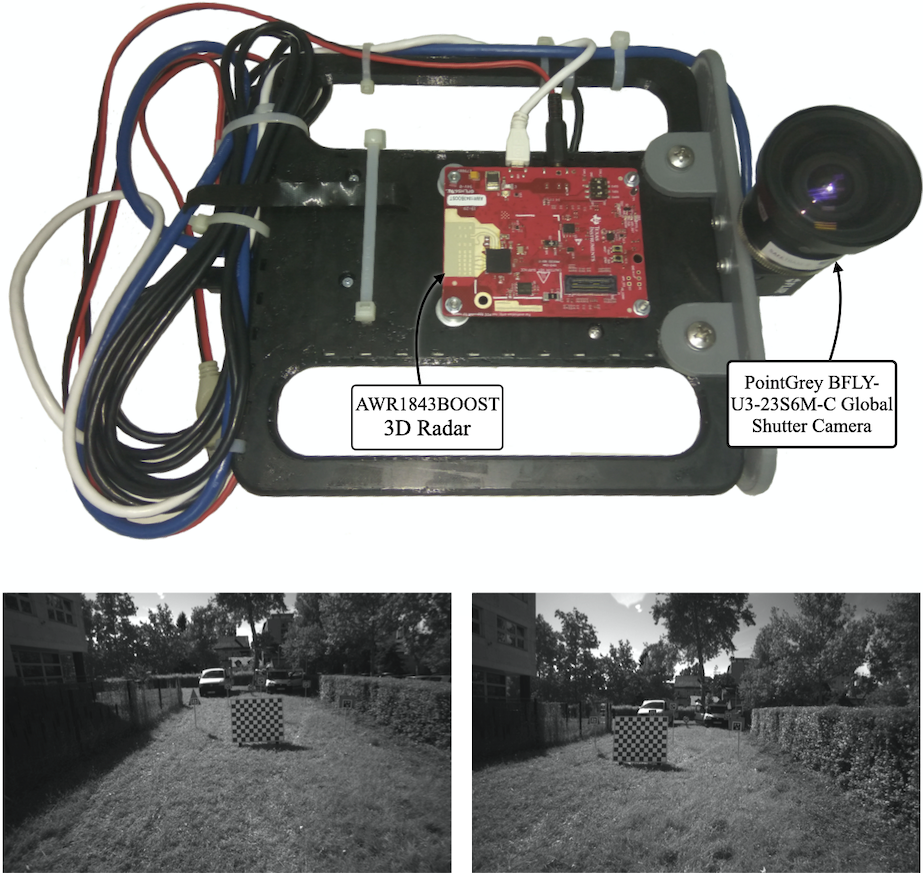}
	\vspace*{-2mm}
	\caption{The top image is a picture of the handheld data collection rig. The bottom two images show different perspectives of our data collection environment.}
	\label{fig:exp_rig}
	\vspace*{-5mm}
\end{figure}

We collected a real-world dataset that allowed us to compare the performance of our algorithm to the 3D reprojection-based algorithm of Per\v{s}i\'{c} et al.\ \cite{persic_spatiotemporal_2019}.
Our data collection rig (shown in Figure \ref{fig:exp_rig}) carried: (i) a PointGrey BFLY-U3-23S6M-C global shutter camera with a Kowa C-Mount 6 mm fixed-focus lens ($96.8^{\circ} \times 79.4{^\circ}$ field of view) and (ii) a Texas Instruments AWR1843BOOST 3D radar unit.
Both sensors operated at approximately 10 Hz.
Data were captured and stored by an on-board Raspberry Pi 4 Model B.
The camera intrinsic and lens distortion parameters were obtained using the Kalibr toolbox \cite{maye_calibration_2013} prior to conducting the experiments.
We performed a rough, ad hoc temporal alignment of the radar and camera data before running our optimization algorithm.
Additionally, the extrinsic calibration (translation and rotation) parameters were carefully measured by hand for comparison. 

Experiments were conducted outdoors to mitigate (to some extent) radar multipath reflections and other detrimental effects.
We placed five specialized hybrid radar-camera targets \cite{Persic2019_calibration} in the environment for validation purposes and for comparison with the calibration method in \cite{persic_spatiotemporal_2019}.
However, we emphasize that our algorithm does not specifically make use of the retroreflective radar targets; the velocity of the radar can be determined independently.

\begin{figure}[b!]
	\vspace{-0.3cm}
	\centering
	\includegraphics[scale = 0.70]{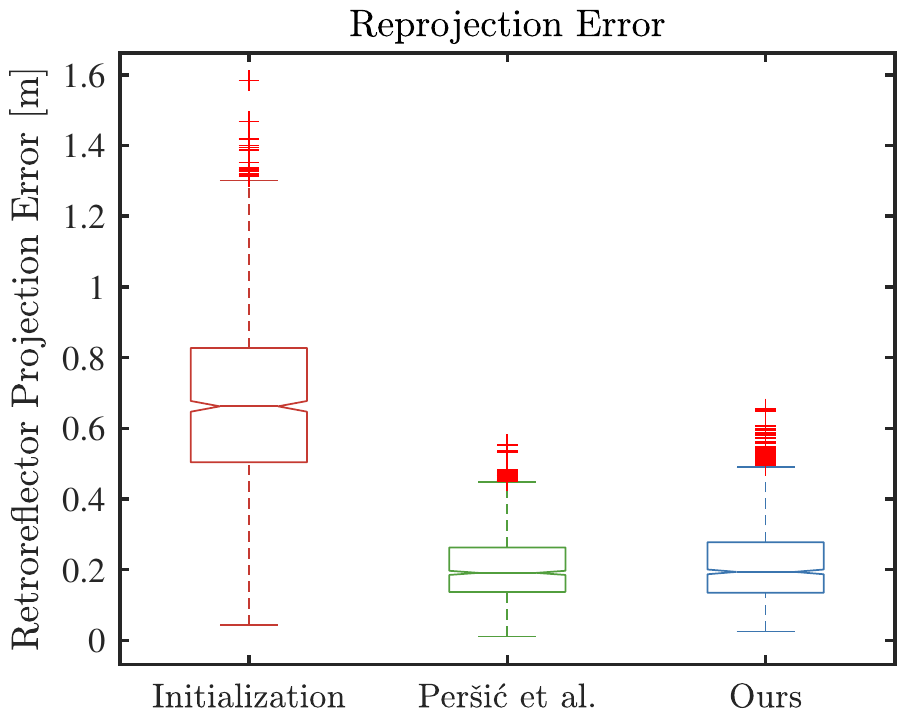}  
	\vspace{-2mm}
	\caption{The target reprojection error is shown for the following calibration methods: hand-measured, Per\u{s}i\'{c} et al.\ \cite{persic_spatiotemporal_2019}, and our proposed method. All algorithms used the same dataset and all calibration results were obtained from a held-out dataset.}
	\label{fig:proj_result}
\end{figure}

We evaluated the performance of the calibration algorithm by measuring target reprojection error.
We placed an AprilTag \cite{olson_tags_2011} on each radar-camera target in the environment, enabling us to estimate the 3D positions of the targets.
Using the extrinsic transform obtained via a given calibration method, the radar measurement of the target can be projected into the camera reference frame.
The distance between the observed 3D position of the target (from image data) and the projected radar estimate of the target position is the target reprojection error.
Figure \ref{fig:proj_result} shows the radar-to-camera reprojection error determined using three different calibration methods: hand-measurement, the 3D reprojection-based method of Per\u{s}i\'{c} et al.\ \cite{persic_spatiotemporal_2019}, and our proposed method.
Since the transform estimated by the 3D reprojection method in \cite{persic_spatiotemporal_2019} optimally aligns the AprilTag positions with the projected radar measurements of the targets, this approach outperforms our algorithm according to this metric, as expected. 
However, the difference in the median reprojection error between our proposed method and that in \cite{persic_spatiotemporal_2019} is less than $4$ mm.
In contrast to \cite{persic_spatiotemporal_2019}, our algorithm does not require any specialized radar targets in the general case.
 
\section{Conclusion}
\label{sec:conclusion}

In this paper, we described a novel continuous-time 3D millimetre-wavelength radar-to-camera extrinsic calibration algorithm.
We showed that the problem is observable and derived the necessary conditions for calibration from radar velocity and camera pose measurements only. 
On synthetic data, our algorithm was shown to be accurate and reliable, but our sensitivity analysis indicated that performance depends on the amount of noise in the radar velocity measurements.
Using data from a handheld sensor rig, we demonstrated that we are able to calibrate the extrinsic transform with an accuracy comparable to the method in \cite{persic_spatiotemporal_2019} but without the need for retroreflectors.
One future research direction is to investigate alternative cost functions that explicitly consider alignment errors (similar to \cite{persic_spatiotemporal_2019}).
Finally, joint spatiotemporal calibration \cite{furgale_unified_2013} and monocular camera trajectory scale estimation, similar to \cite{wise_handeye_2020}, would be valuable extensions to our algorithm.

\bibliographystyle{IEEEtran}
\bibliography{radar-calibration} 

\end{document}